\documentclass[10pt,final,journal,twocolumn]{IEEEtran}
\usepackage{amsmath}
\usepackage{graphicx, epstopdf}
\usepackage{setspace}
\usepackage{acronym}
\usepackage{commath}
\usepackage{mathrsfs}
\usepackage{mathtools}
\usepackage{calc}
\usepackage[top=1in, left=0.7in, right=0.7in]{geometry}
\usepackage{ifthen}
\usepackage{textcomp}
\usepackage{float}
\usepackage[tight,footnotesize]{subfigure}
\usepackage{cite}
\usepackage{bm}
\usepackage{url}
\usepackage[utf8]{inputenc}
\usepackage{mathtools}
\usepackage{amssymb}
\usepackage{amsthm}
\usepackage{mathrsfs}
\usepackage{bm}
\usepackage{multirow}
\usepackage{multicol}
\usepackage{xspace}
\usepackage{tabularx}

\usepackage{cite}
\usepackage{amsmath,amssymb,amsfonts}
\usepackage{algorithmic}
\usepackage{algorithm}
\usepackage{graphicx}
\usepackage{textcomp}

\usepackage{xcolor}
\def\BibTeX{{\rm B\kern-.05em{\sc i\kern-.025em b}\kern-.08em
    T\kern-.1667em\lower.7ex\hbox{E}\kern-.125emX}}

\begin{document}

\title{AI Evasion and Impersonation Attacks on Facial Re-Identification with Activation Map Explanations}

\author{Noe Claudel$^{1}$, Weisi Guo$^{1*}$, Yang Xing$^{1}$
\thanks{$^{1}$All authors are with the Centre for Assured and Connected Autonomy, Cranfield University, Bedford, UK. $^{*}$Corresponding author: weisi.guo@cranfield.ac.uk. 
This work is supported by Bedfordshire Police.}
}

\maketitle

\begin{abstract}
Facial identification systems are increasingly deployed in surveillance and security applications, yet their vulnerability to adversarial evasion attacks pose a critical risk. This paper introduces a novel framework for generating adversarial patches capable of performing both evasion and impersonation attacks against deep re-identification models across non-overlapping cameras. Unlike prior approaches that require iterative optimization for each target, our method employs a conditional encoder–decoder network to synthesize adversarial patches in a single forward pass, guided by multi-scale features from source and target images. The patches are optimized with a dual adversarial objective comprising pull and push terms. To enhance imperceptibility and aid physical deployment, we further integrate naturalistic patch generation using pre-trained latent diffusion models. 

Experiments on standard pedestrian (Market-1501, DukeMTMCreID) and facial recognition benchmarks (CelebA-HQ, PubFig) datasets demonstrate the effectiveness of the proposed method. Our adversarial evasion attacks reduce mean Average Precision from 90\% to 0.4\% in white-box settings and from 72\% to 0.4\% in black-box settings, showing strong cross-model generalization. In targeted impersonation attacks, our framework achieves an success rate of 27\% on CelebA-HQ, competing with other patch-based methods. We go further to use clustering of activation maps to interpret which features are most used by adversarial attacks and propose a pathway for future countermeasures. The results highlight the practicality of adversarial patch attacks on retrieval based systems and underline the urgent need for robust defense strategies.  
\end{abstract}


\section{Introduction}
Person re-identification (Re-ID) is the problem of retrieving images of the same individual captured by non-overlapping cameras. In other words, given a query person image the system must determine if that person appears in other camera views. Re-ID is widely used in security, surveillance and is challenging because must contend with variations in viewpoint, illumination, resolution, and background content. In recent years, the development of smart city technologies has enhanced urban living by integrating this advanced system that improve safety and efficiency. Modern systems use deep learning which achieve state-of-the-art performance in person re-identification. However, the robustness of deep neural networks in these contexts has become a key concern, as numerous studies have revealed their vulnerability to adversarial attacks. 

An adversarial attack can be categorized along several dimensions \cite{10284006, 10268441}:
\begin{itemize}
    \item White-box vs Black-box attack: In white box attacks the adversary knows the model parameters and architecture, whereas in black-box attacks they do not. 
    \item Training vs Inference: Poisoning attacks inject malicious examples into the training data using backdoor access, while evasion attacks perturb inputs at inference time.
    \item Digital vs Physical: Digital attacks are applied directly to pixel inputs, whereas physical attacks \cite{10233999} assume the perturbation is printed or worn in the real world. 
    \item Targeted vs Untargeted: In impersonation (targeted) attack, the goal is to have the person recognized as a specific other identity while in evading attack (untargeted) the goal is simply to avoid correct recognition.
    \item Conspicuous vs Unconspicous: Conspicuous attacks are visually obvious and may raise suspicion. In contrast unconspicuous attack are designed to blend into the natural appearance of the scene or object \cite{8854834, pmlr-v97-li19j}.
\end{itemize}

\subsection{Review on Evasion Attacks}
The first physical adversarial attack for Re-ID was proposed by Wang et al. \cite{Wang_2019_ICCV} called AdvPattern. They generated printable patches to evade search or impersonate a target identity. Following this, Ding et al. \cite{9432915} introduced a universal adversarial perturbation (MUAP) that disrupts the similarity ranking of a Re-ID model. This is a digital attack that showed that a single perturbation pattern added to any query image can significantly degrade retrieval accuracy. Wang et al. \cite{Wang_2020_CVPR} similarly attacked the similarity computation to induce incorrect rankings. Wang et al. \cite{10.1145/3503161.3547958} proposed using adversarial attacks as a form of target protection crafting perturbation, so the person’s feature does not match any gallery identity. Gong et al. \cite{gong2024cross} extended universal perturbations to visible-infrared cross-modality Re-ID, though with limited real-world applicability. A landmark result by Sharif et al. \cite{sharif2016accessorize} demonstrated that printed adversarial eyeglass frames can cause high mis-recognition rates. They optimized a perturbation pattern on eyeglass and achieved up to 97\% success in dodging attacks and 75\% in impersonation attacks. In 3-dimensional or soft body cases, stretch invariance can be added to the patch training process to take into account real-world rotation and stretch transformations \cite{10233999}. In summary, extensive prior work has shown that deep Re-ID and face recognition models can be misled by adversarial patches. All the methods cited above help motivate our approach and establish performances baselines for comparison.

\subsection{Gap and Novelties}
The current gaps in research include 2 areas: (1) the ability for attacks to mis-identify or impersonate to a different target especially in black-box settings, and (2) the practical deployability of these patch attacks under diverse settings such as single training process, printer friendly, and low human perceptibility (unconspicous). We propose a pose-invariant adversarial patch generation framework designed to mislead deep learning-based re-identification system. Our novelties are:
\begin{enumerate}
    \item One-Off Targeted Attack Training: developed targeted and untargeted attack in \textit{one forward pass} of a patch generator network without new training for each new identity targeted. The method integrates conditional encoder-decoder network that synthesizes patches informed by multi-scale features from both a source and a target image.
    \item Improved Imperceivable (Unconspicous) and Deployment Attributes: To improve the stealth and printability of the adversarial patches we further introduce a naturalistic patch generation strategy. We introduce the perturbation in the latent space of a pretrained diffusion model to preserve realism while maintaining adversarial effectiveness. 
\end{enumerate}
Our innovations are rigorously evaluated on two pedestrian re-identification benchmarks (Market1501 and DukeMTMC) as well as two facial recognition datasets (PubFig and CelebA-HQ) under both white-box and black-box attack assumptions.

\begin{figure}[!t]
\centering
\includegraphics[width=3.3in]{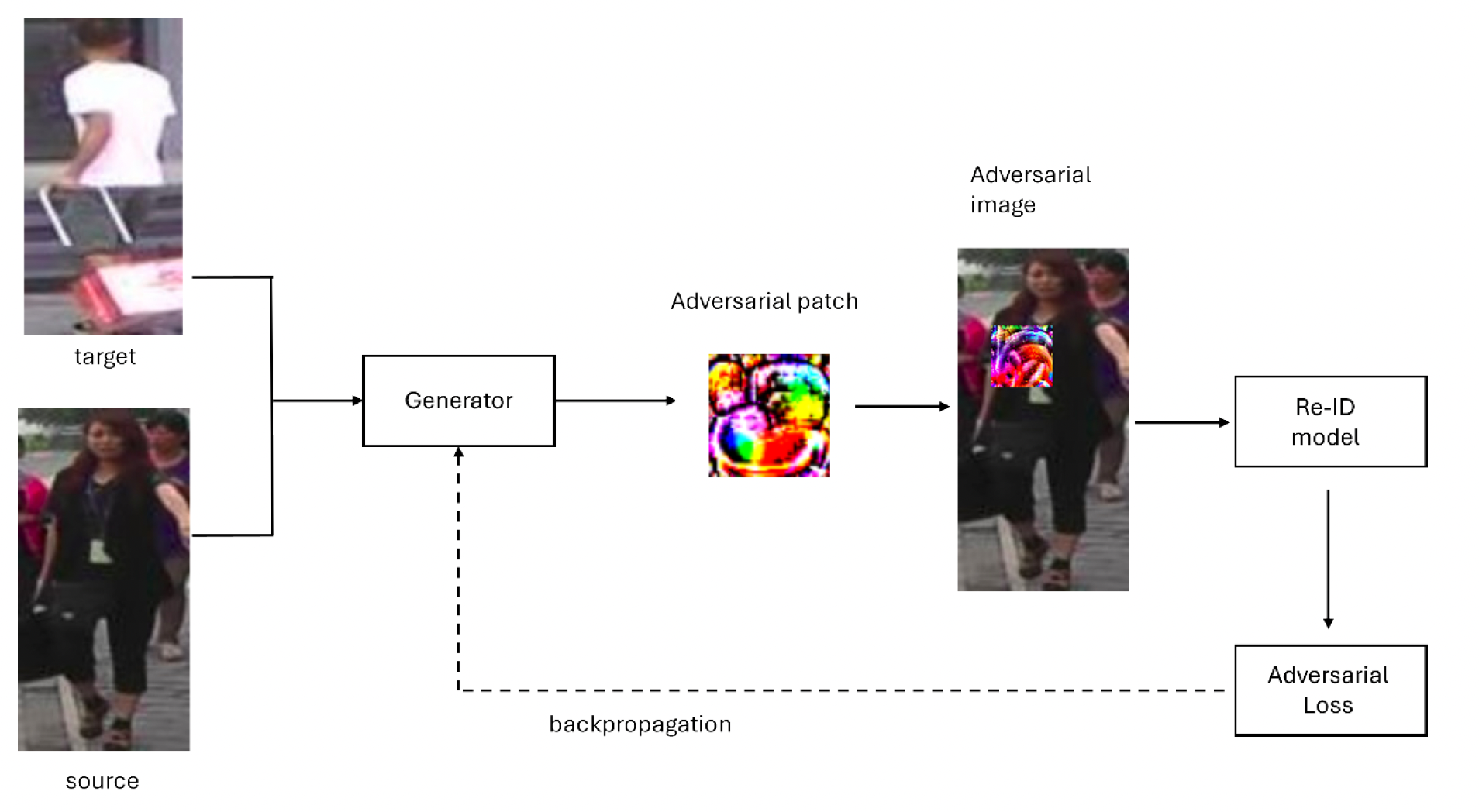}
\caption{Architectural overview of adversarial patch generation workflow.}
\label{fig1}
\end{figure}

\section{System Setup and Methodology}

Let $I_s \in \mathbb{R}^{H \times W \times 3}$ denote a source (probe) image containing a person whose identity is $y_s$, and let $I_t \in \mathbb{R}^{H \times W \times 3}$ denote a target gallery image of a different persons with identity $y_t$. A trained ReID model $f$ maps each input image to a feature embedding $f(I) \in \mathbb{R}^{d}$. The ReID decision relies on the distance (e.g., Euclidean or cosine) between probe and gallery embeddings. Our goal is to generate a sparse, learnable patch $P \in \mathbb{R}^{h \times w \times 3}$ such that, when $P$ is pasted onto $I_s$, the resulting adversarial image:
\begin{equation}
    I_{adv} = (1-M) \oplus I_s + M \oplus T(P),
\end{equation}
fools the ReID model into matching $I_{adv}$ with the targeted attack $I_{t}$ while moving it away from $I_{s}$. Here $T$, denotes a set of random spatial and photometric
transformations applied to $P$ (e.g., affine warp, perspective, Gaussian blur) to
increase robustness.

During the training phase, one or more pre-trained person re-identification or facial recognition models, called targeted models, are employed to extract feature embeddings from the adversarially perturbed images. These embeddings represent the features that the targeted model uses for matching. The generator network which produces the patch is optimized so that when the patch is applied to an image, the resulting embedding leads to a successful attack.

A global overview of the method is shown on Figure \ref{fig1}. To achieve this the extracted embeddings are used to compute the adversarial loss function, which directly constitute the training objective. During the inference phase, the process is more lightweight. Only the trained generator is required to produce the adversarial patch. Given a source image and, if the scenario is targeted, a target image representing the desired identity, the generator creates the adversarial patch.

\subsection{Patch Generation}
This design embeds target‐specific pedestrian attributes into the patch, enhancing its ability to mimic gallery embeddings. We design a conditional, encoder–decoder network that synthesizes $P$ based on
multi‐scale features from $I_{s}$ both and $I_{t}$:
\begin{itemize}
    \item Feature Extraction: A pretrained backbone (e.g., ResNet‐50) encodes each image to extract deep feature maps at three spatial resolutions for source and target.
    \item Feature Fusion: We concatenate the highest‐level source and target features along the channel dimension and reduce them to 512 channels via a 1×1 convolution.
    \item Up-sampling Decoder: Four sequential Up-Blocks progressively up-sample the fused feature to the patch spatial size ($h, w$). At each stage, skip connections inject corresponding target features to guide shape and texture synthesis. The Up-Blocks in the generator architecture are designed to progressively increase the spatial resolution while integrating information from the target features at each stage. Each block consists of a transpose convolution layer that up samples the spatial dimensions of the input feature map by a factor 2. Feature concatenation where the up-sampled feature map is concatenated with the corresponding target feature map. Finally, a 2d convolution layer processes the concatenated features to refine them and learn effective joint representation. This convolution is followed by a ReLU activation and by a batch normalization.
    \item Output Head: A final 3×3 convolution followed by a tanh activation produces the RGB patch $P$, whose values are scaled to [―1,1].
\end{itemize}

\subsection{Patch Embedding and Differential Blending}
To paste the patch onto $I_s$, we sample a random location $(x,y)$. within the image boundaries and generate a binary mask $M$ of the same size as $I_s$, where values inside a $h \times w$ window at $(x,y)$ are 1, and 0 elsewhere. We then blend:
\begin{equation}
    I_{adv} = (1-M) \oplus I_s + M \oplus P_{tiled},
\end{equation}
where $P_{tiled}$ is the patch broadcast to the full image size. This affine blending remains fully differentiable with respect to the $P$.

\subsection{Adversarial Objectives}
\subsubsection{Targeted Misidentification Attack}
Using the ReID model $f$ as a fixed feature extractor, we denote the normalized embeddings of $I_s, I_t, I_{adv}$ and $z_s, z_t, z_{adv} \in \mathbb{R}^{d}$. Our loss function is comprised of a push and pull term augmented by hyperparameters: 
\begin{equation}
    \mathcal{L}_{adv} = \lambda_1 \mathcal{L}_{pull} + \lambda_2 \mathcal{L}_{push},
\end{equation} 

The \textbf{pull term} is:
\begin{equation}
    \mathcal{L}_{pull} = 1-\cos{(z_{adv}, z_t)},
\end{equation} 
which encourages $I_{adv}$ to move closer to the target via the cosine similarity function.

The \textbf{push term} is:
\begin{equation}
    \mathcal{L}_{push} = \max (0, 1-\cos{(f_{adv}, f_t)}) + \tau - [1-\cos{(f_{adv}, f_s)}],
\end{equation} where $\tau$ penalizes the cosine similarity function to drive $I_{adv}$ away from the original identity. By introducing the margin-based formulation, the push effect is only applied when the similarity to the source identity is above $\tau$. Once the similarity drops below the margin the penalty becomes zero. This ensures a better balance between moving away from the original identity and converging toward the target identity.

\subsubsection{Untargeted Evasion Attack}
We only seek to evade the source identity, without pushing the embedding toward any specific target. Here, we only define a single “push” term that penalizes high similarity between $I_{adv}$, and its identity: 
\begin{equation}
    \mathcal{L}_{push} = \cos{(z_{adv}, z_s)},
\end{equation} where by only separating it from its original identity, we simplify the loss landscape and achieve faster convergence and stronger performance. This is the baseline attack setting which is a derivative of the targeted attack.

\begin{figure}[!t]
\centering
\includegraphics[width=3.3in]{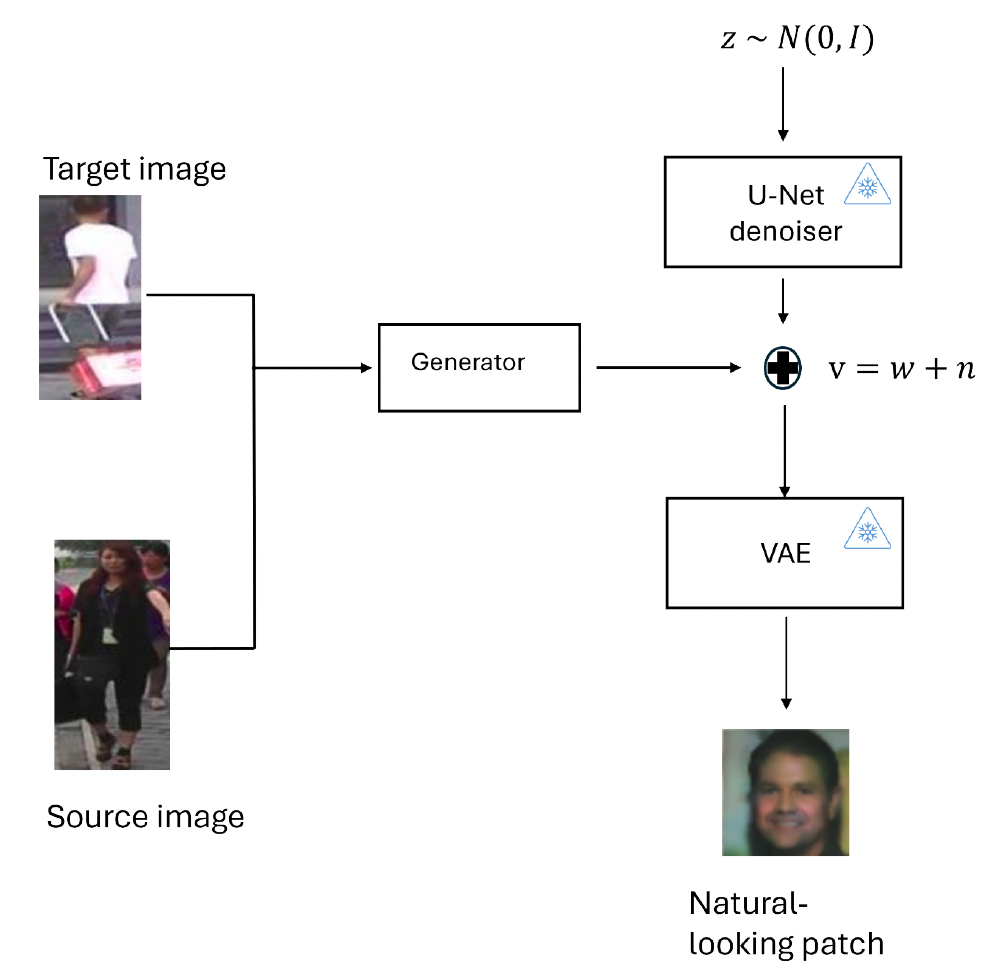}
\caption{Pipeline of natural patch generation using latent space perturbations.}
\label{fig2}
\end{figure}

\section{Training Procedure and Naturalistic Synthesis}
We optimize the patch generator parameters $\theta$ via stochastic gradient descent
over mini batches of image pairs $I_s, I_t$ and we train for 100 epochs with an initial learning rate of $2x10^{-4}$, $\tau = 0.3$, $\lambda_1 = 1, \lambda_2 = 0.5$. During each iteration, we sample a random location and transformation for each
patch, ensuring varied adversarial examples. The resulting generated patch has an aggressive texture which make it hard to print. Even though there are many tricks to constraints the pixel values distribution such as the TV loss it is still very conspicuous and might raise attentions. So that leveraging image generative model can help to create natural looking adversarial patches. We focused on two different try to get natural looking image:
\begin{enumerate}
    \item GAN-based Adversarial Patch Synthesis: During training, the discriminator parameters are updated by minimizing: $\mathcal{L}_{D} = - [E_{x \sim p_{data}} (\log_D x) + E_{z \sim p_{z}} (\log (1-D(G(z))))]$, where we penalize the first term $D$ if it assigns low realness to patches . Simultaneously the generator is trained to both fool the discriminator and to produce a patch when overlaid on inputs, causes mis-classification in the target model $f$, with loss function: $\mathcal{L}_{G} = - E_{z \sim p_{z}} [\log (D(G(z)))] + \lambda_{adv}\mathcal{L}_{adv}$.
    \item Latent Space Perturbation in a Pretrained Diffusion Model: The second approach exploits the exceptionally high image quality of pretrained latent diffusion models \cite{rombach2022high}. A diffusion pipeline consists of a denoising network $f_{denoise}$ that iteratively refines a noise tensor $I_{noisy}$ into a clean latent code $l \in \mathbb{R}^{C\times h \times w}$, and a Variational Auto-Encoder (VAE) decoder that reconstructs pixel-space images from $l$. To achieve low practical imperceptibility, we introduce a learnable perturbation $\delta \in \mathbb{R}^{C\times h \times w}$ into the denoised latent space: $\overline{l} = l + \delta$, and pass it through the VAE to obtain the final image. Optimising for $\delta$ needs to regularise the perturbations by penalizing deviations in latent norm.
\end{enumerate}

Figure\ref{fig2} illustrates the overall workflow of the natural patch generation process. The adversarial generator produces an output vector that is injected directly into the latent space of the LDM by adding it to the original latent representation of the image. This perturbed latent representation is then decoded through the LDM’s VAE to obtain the final adversarial patch in the image space. Importantly, both the U-Net denoiser and the VAE decoder of the LDM remain frozen throughout adversarial training, ensuring that only the generator is updated. This design choice prevents degradation of the generative prior and guarantees that the adversarial patches retain the naturalness and realism imposed by the diffusion model, while still being optimized to fool the downstream Re-ID or face recognition system.

\section{Data Sets, Models, and Assumptions}
To train and evaluate our adversarial patch generator in the person re-identification (ReID) setting, we rely on two standard public benchmarks widely used in the literature: The market-1501 and the DukeMTMC-reID. The first has a total of 32 668 annotated pedestrian images captured by 6 cameras. It contains 1 501 unique persons (751 for training, 750 for testing). DukeMTMC-reID has 36 411 images from 8 cameras and it contains 1 812 identities (702 training, 702 testing, 408 distractors).In the evading attack scenario we consider, the objective is to prevent the model from correctly matching a person’s appearance across different camera views. That is, the adversarial patch is optimized to cause mismatches between different images of the same identity, breaking the continuity of tracking and recognition across a camera network. These two datasets are particularly well-suited for this task, as they include multiple images per person captured under varying viewpoints and environmental conditions.

The target model refers to the re-identification network used during the training phase of the adversarial patch generator. In our setup, this model is based on OSNet, a lightweight convolutional architecture specifically designed for person re-identification. We train OSNet on the Market-1501 dataset using the combined standard identity classification objective and the triplet loss, allowing it to learn robust feature embeddings for distinguishing individuals across camera views. To assess the transferability of the adversarial patch and its general effectiveness beyond a single model, we introduce an auxiliary model for evaluation. This auxiliary network is a ResNet-50, also trained on the Market-1501 dataset using identical supervision. Importantly, the ResNet-50 model is not exposed to the patch generator during training. It thus serves as an independent benchmark to evaluate whether the evading attack generalizes to unseen model architectures. This setup enables us to measure both:
\begin{itemize}
    \item White-box performance, where the adversarial patch attacks the model, it was trained against (OSNet).
    \item Black-box performance, where the patch is evaluated on a structurally different, independently trained model (ResNet-50).
\end{itemize}
By comparing results across both models, we gain insight into the robustness and generalization ability of our adversarial evasion strategy.

\section{Results}
\begin{figure}[!t]
\centering
\includegraphics[width=3.3in]{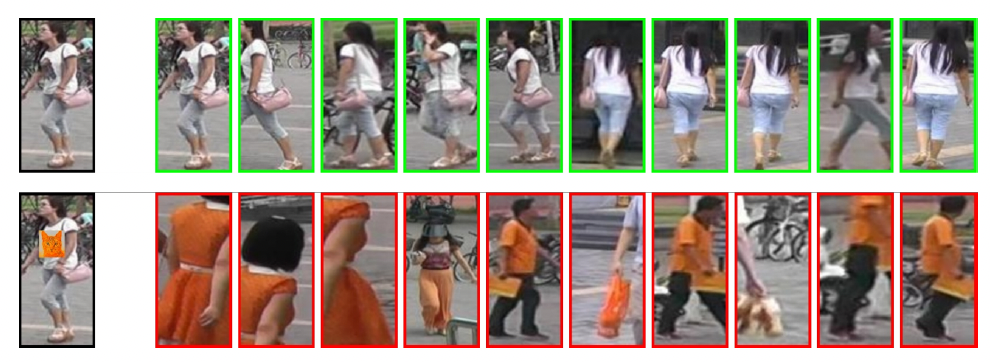}
\caption{Top 10 retrieval for the same probe non-attacked and attacked.}
\label{fig3}
\end{figure}

Figure \ref{fig3} illustrates the effect of the adversarial evasion attack using a natural looking patch. The figure compares the top 10 retrieval results for the same probe image in two settings: without and with the adversarial patch. Without attack, the system correctly retrieves images corresponding to the same identity. In contrast, once adversarial patch is applied none of the top 10 retrieved images correspond to the correct identity. This highlights the ability of the naturalistic adversarial patch to mislead the recognition system.

\subsection{Evasion Attacks}
The white and black box assumption results in Table 1 and 2 mean the attack either understands the classifier used or does not. To quantitatively assess the effectiveness of the adversarial evading attack, we adopt the mean Average Precision (mAP) metric, which is the standard benchmark for retrieval-based tasks. A lower mAP after patch application indicates that the model struggles to recognize the adversarially perturbed person, confirming the effectiveness of the attack. 

The white box mAP performances show our adversarial patch can erode mAP from 90\% to 0.4\% or 5\% if we apply a naturalistic patch. The other panel results show the erosion of 10th likely match and 1st likely match, which follows a similar pattern. In the black box case, the naturalistic patch achieves a far less desirable performance, as the naturalistic constraint limits the expressiveness of the patch restricting the extent to which adversarial features can be encoded and transferred to other unknown black box models. Other variables we examined:
\begin{itemize}
    \item Size of Patch: as the size of the patch decreases, the mAP can increase from 0.3\% to 60\% for a patch size decrease of 3x.
    \item Data Set Transferability: we tried the DukeMTMC-reID data set and our results still achieved 0.3\% mAP.
\end{itemize}

\begin{table}[ht]
    \centering
    \caption{Evasion Performances under white box setting}
    \label{tab_1}
    \begin{tabular}{lccc}
    \hline
                    & mAP & 10th & 1st \\
    \hline
    No Patch        & 90\% & 100\% & 100\%  \\
    Random Patch    & 75\% & 94\% & 98\%  \\
    Adversarial Patch & 0.4\% & 0.3\% & 0.9\%  \\
    Naturalistic Patch & 5\% & 3.4\% & 9.2\%  \\
    \end{tabular}
\end{table}

\begin{table}[ht]
    \centering
    \caption{Evasion Performances under black box setting}
    \label{tab_2}
    \begin{tabular}{lccc}
    \hline
                    & mAP & 10th & 1st \\
    \hline
    No Patch        & 72\% & 89\% & 97\%  \\
    Random Patch    & 68\% & 87\% & 96\%  \\
    Adversarial Patch & 0.4\% & 0.3\% & 5\%  \\
    Naturalistic Patch & 43\% & 61\% & 87\%  \\
    \end{tabular}
\end{table}

\begin{figure}[!t]
\centering
\includegraphics[width=3.3in]{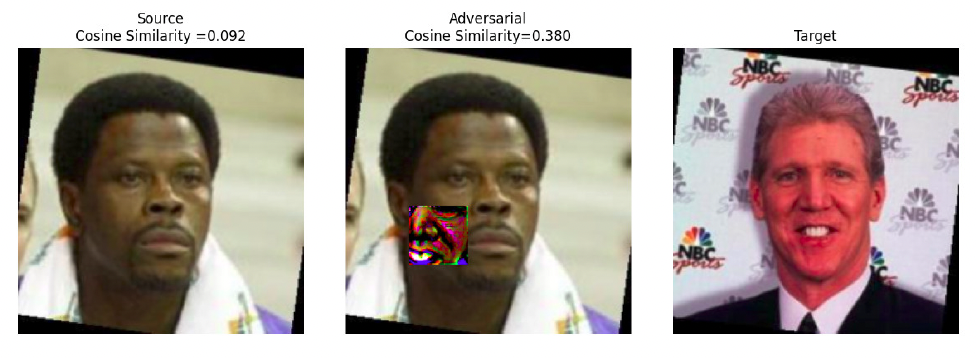}
\caption{Example of impersonation attack.}
\label{fig4}
\end{figure}

\begin{figure*}[!t]
\centering
\includegraphics[width=7in]{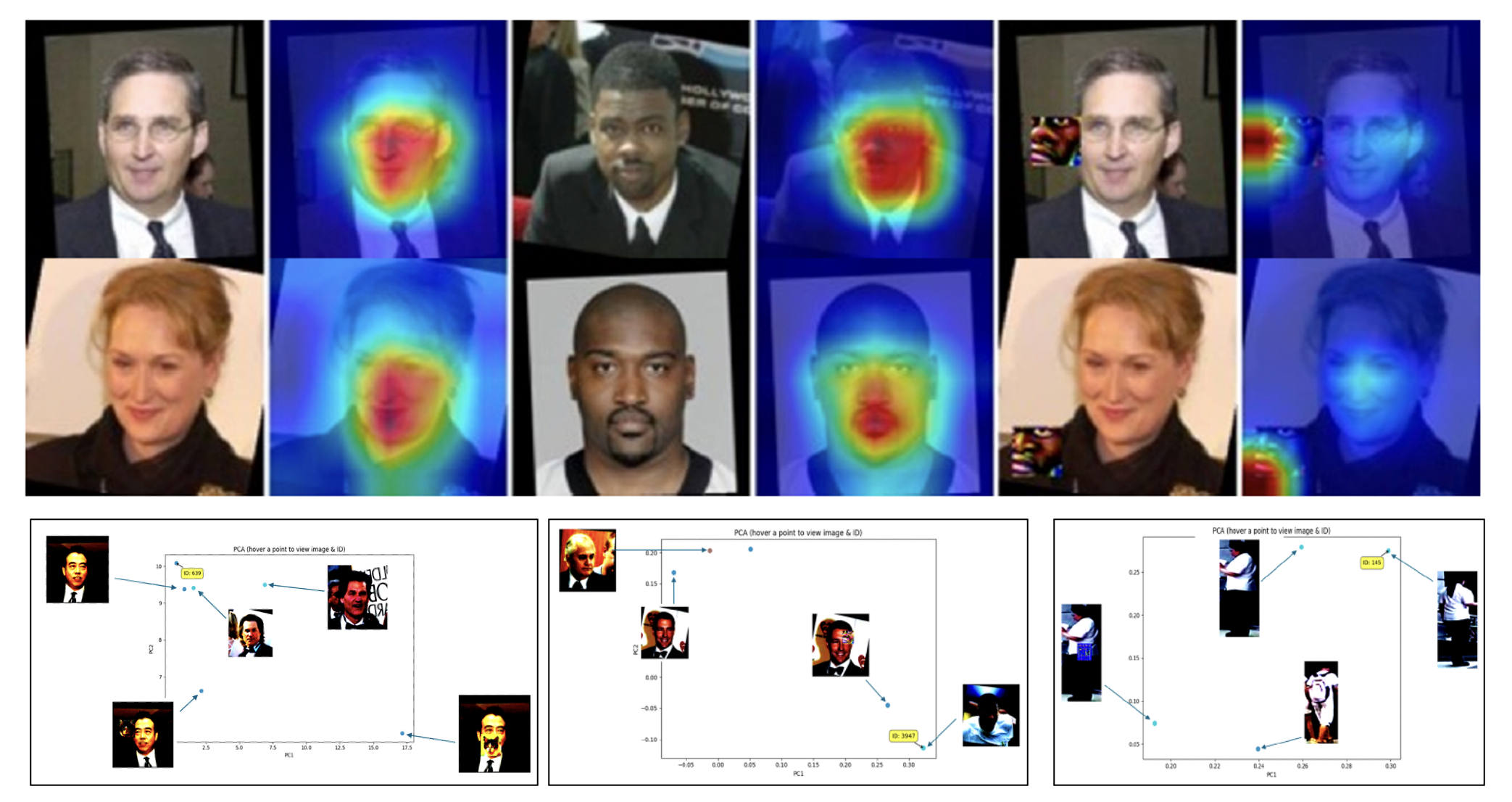}
\caption{(Top) Examples of activation maps highlighting the image regions most
activated by the target model. (Bottom) Embeddings projected in 2D space in context of evasion attack on face recognition system.}
\label{fig5}
\end{figure*}

\subsection{Targeted Mis-classification Attacks}
To quantitatively assess the effectiveness of the adversarial mis-classification target attack, we adopt the attack success rate (ASR) metric, which is the proportion of adversarial examples for which the similarity with the attacker’s target identity exceeds a pre-defined threshold $\tau$:
\begin{equation}
    ASR = \frac{1}{N} \sum^{N}_{i=1} \cos{(f_{adv}^{i},f_{t}^{i}) > \tau},
\end{equation} where $N$ is the number of trials and a high ASR indicates that more patches successfully deceive the model into matching the attacker’s target. Overall, we achieve a 54\% ASR with ResNet50 and 27\% ASR with FaceNet in attack success. A detailed breakdown is given in Table 3, and comparative performance is shown in Table 4. Whilst our approach is not better in terms of performance, we only require a single training process.

\begin{table}[ht]
    \centering
    \caption{Impersonation Cosine Similarity Performances}
    \label{tab_3}
    \begin{tabular}{lcccc}
    \hline
                            & source & target & random & accuracy \\
    \hline
    ResNet50 No Patch        & 85\% & 66\% & 68\%  & 92\% \\
    ResNet50 with Patch      & 76\% & 77\% & 66\%  & 47\% \\
    FaceNet No Patch         & 75\% & 18\% & 20\%  & 98\% \\
    FaceNet with Patch       & 69\% & 48\% & 22\%  & 76\% \\
    \end{tabular}
\end{table}

\begin{table}[ht]
    \centering
    \caption{Impersonation Attack Benchmark Against State-of-the-Art}
    \label{tab_4}
    \begin{tabular}{lc}
    \hline
                                  & Performance \\
    \hline
    Adv-Hat \cite{9412236} & 4.7\% \\
    Adv-Glasses \cite{sharif2016accessorize}    & 9.1\% \\
    Our Approach    & 26.7\% \\
    \end{tabular}
\end{table}

\section{Activation Map Explanations}
To better understand what features the patch generator extracts and utilizes to create adversarial patches, we propose a visualization of the activation maps from the last feature layer of the feature extractor. This analysis provides insight into which regions of the input the model focuses on during patch generation. Given an activation tensor $F \in \mathbb{R}^{B\times C\times h \times w}$ we compute a 2D activation map per image by summing the squared activations across the channel dimensions: $A_{i,j} = \sum_{c=1}^{C} F^{2}_{c,i,j}$. This results in a single 2D activation map per image. The visualization of the activation maps in Figure\ref{fig5} top shows how the patch influences the targeted model’s internal representations. This visualization reveals that the patch draws and concentrates nearly all of the model’s attention, effectively dominating the feature representation and overshadowing other facial regions.

We then employ Principal Component Analysis (PCA) as a visual and quantitative tool to assess the adversarial effect on the generated patch on image embeddings. The core idea is to compare the position of attacked sample in the feature space to that of their clean and target counterparts. The procedure is as follows: The L2-normalised embeddings are extracted from the targeted model for: (a) source image, (b) clean comparison image of same source, (c) a target image, and (d) the adversarial patched image. The PCA is trained on features matrix $X \in \mathbb{R}^{N \times d}$. We analysed the embeddings’ projections under different attack scenarios. Figure\ref{fig5} bottom illustrates an evasion attack on face recognition using natural-looking generated patches: the attacked images are pushed far from their original source points and/or drawn toward the target.

\section{Conclusions and Future Work}
The key novelty of our approach lies in the ability to generate adversarial patches without retraining for each new target identity, making it more efficient and adaptable than prior works. The promising results for both evading and targeted attacks suggest that our architecture provides a flexible framework for exploring adversarial vulnerabilities in re-id systems. 

Our evading attack experiments demonstrate strong effectiveness by reducing the mAP of the Re-ID model to below 1\%. The generated patches show good transferability across different models and datasets, highlighting the robustness of our method. By incorporating a pre-trained latent diffusion model, we were also able to produce perturbations that retain natural-looking appearance enhancing the stealthiness of the adversarial examples. 

For the targeted attack, we evaluated the attack success rate on the CelebA-HQ dataset and found that our method achieves a slightly lower performance compared to existing state-of-the-art without having to be re-trained for different targets which is a crucial advantage. 

Future work should further investigate real-world physical deployment, evaluate targeted attacks on large-scale re-id datasets, and explore adaptive defences that can mitigate the transferability of our generated patches.

\bibliographystyle{IEEEtran}
\bibliography{main.bib}

@ARTICLE{10268441,
  author={Guesmi, Amira and Hanif, Muhammad Abdullah and Ouni, Bassem and Shafique, Muhammad},
  journal={IEEE Access}, 
  title={Physical Adversarial Attacks for Camera-Based Smart Systems: Current Trends, Categorization, Applications, Research Challenges, and Future Outlook}, 
  year={2023},
  volume={11},
  number={},
  pages={109617-109668},
}

@ARTICLE{10284006,
  author={Nguyen, Kien and Fernando, Tharindu and Fookes, Clinton and Sridharan, Sridha},
  journal={IEEE Transactions on Neural Networks and Learning Systems}, 
  title={Physical Adversarial Attacks for Surveillance: A Survey}, 
  year={2024},
  volume={35},
  number={12},
  pages={17036-17056},
  doi={10.1109/TNNLS.2023.3321432}}

@InProceedings{Wang_2019_ICCV,
author = {Wang, Zhibo and Zheng, Siyan and Song, Mengkai and Wang, Qian and Rahimpour, Alireza and Qi, Hairong},
title = {advPattern: Physical-World Attacks on Deep Person Re-Identification via Adversarially Transformable Patterns},
booktitle = {Proceedings of the IEEE/CVF International Conference on Computer Vision (ICCV)},
month = {October},
year = {2019}
}

@InProceedings{Wang_2020_CVPR,
author = {Wang, Hongjun and Wang, Guangrun and Li, Ya and Zhang, Dongyu and Lin, Liang},
title = {Transferable, Controllable, and Inconspicuous Adversarial Attacks on Person Re-identification With Deep Mis-Ranking},
booktitle = {Proceedings of the IEEE/CVF Conference on Computer Vision and Pattern Recognition (CVPR)},
month = {June},
year = {2020}
}

@ARTICLE{9432915,
  author={Ding, Wenjie and Wei, Xing and Ji, Rongrong and Hong, Xiaopeng and Tian, Qi and Gong, Yihong},
  journal={IEEE Transactions on Information Forensics and Security}, 
  title={Beyond Universal Person Re-Identification Attack}, 
  year={2021},
  volume={16},
  number={},
  pages={3442-3455},
  doi={10.1109/TIFS.2021.3081247}}

@INPROCEEDINGS{10233999,
  author={Li, Chen and Guo, Weisi},
  booktitle={2023 IEEE Ninth International Conference on Big Data Computing Service and Applications (BigDataService)}, 
  title={Soft Body Pose-Invariant Evasion Attacks against Deep Learning Human Detection}, 
  year={2023},
  volume={},
  number={},
  pages={155-156},
  doi={10.1109/BigDataService58306.2023.00032}}

@InProceedings{pmlr-v97-li19j,
  title = 	 {Adversarial camera stickers: A physical camera-based attack on deep learning systems},
  author =       {Li, Juncheng and Schmidt, Frank and Kolter, Zico},
  booktitle = 	 {Proceedings of the 36th International Conference on Machine Learning},
  pages = 	 {3896--3904},
  year = 	 {2019},
  editor = 	 {Chaudhuri, Kamalika and Salakhutdinov, Ruslan},
  volume = 	 {97},
  series = 	 {Proceedings of Machine Learning Research},
  month = 	 {09--15 Jun},
  publisher =    {PMLR},
}

@ARTICLE{8854834,
  author={Chen, Chen and Zhao, Xinwei and Stamm, Matthew C.},
  journal={IEEE Transactions on Information Forensics and Security}, 
  title={Generative Adversarial Attacks Against Deep- Learning-Based Camera Model Identification}, 
  year={2025},
  volume={20},
  number={},
  pages={7679-7694},
  doi={10.1109/TIFS.2019.2945198}}

@inproceedings{rombach2022high,
  title={High-resolution image synthesis with latent diffusion models},
  author={Rombach, Robin and Blattmann, Andreas and Lorenz, Dominik and Esser, Patrick and Ommer, Bj{\"o}rn},
  booktitle={Proceedings of the IEEE/CVF conference on computer vision and pattern recognition},
  pages={10684--10695},
  year={2022}
}

@inproceedings{10.1145/3503161.3547958, author = {Wang, Lin and Zhang, Wanqian and Wu, Dayan and Zhu, Fei and Li, Bo}, title = {Attack is the Best Defense: Towards Preemptive-Protection Person Re-Identification}, year = {2022}, isbn = {9781450392037}, publisher = {Association for Computing Machinery}, address = {New York, NY, USA}, url = {https://doi.org/10.1145/3503161.3547958}, doi = {10.1145/3503161.3547958},booktitle = {Proceedings of the 30th ACM International Conference on Multimedia}, pages = {550–559}, numpages = {10}, }

@article{gong2024cross,
  title={Cross-modality perturbation synergy attack for person re-identification},
  author={Gong, Yunpeng and Zhong, Zhun and Qu, Yansong and Luo, Zhiming and Ji, Rongrong and Jiang, Min},
  journal={Advances in Neural Information Processing Systems},
  volume={37},
  pages={23352--23377},
  year={2024}
}

@inproceedings{sharif2016accessorize,
  title={Accessorize to a crime: Real and stealthy attacks on state-of-the-art face recognition},
  author={Sharif, Mahmood and Bhagavatula, Sruti and Bauer, Lujo and Reiter, Michael K},
  booktitle={Proceedings of the 2016 acm sigsac conference on computer and communications security},
  pages={1528--1540},
  year={2016}
}

@INPROCEEDINGS{9412236,
  author={Komkov, Stepan and Petiushko, Aleksandr},
  booktitle={2020 25th International Conference on Pattern Recognition (ICPR)}, 
  title={AdvHat: Real-World Adversarial Attack on ArcFace Face ID System}, 
  year={2021},
  volume={},
  number={},
  pages={819-826},
  doi={10.1109/ICPR48806.2021.9412236}}

\end{document}